\documentclass[lettersize,journal]{IEEEtran}
\usepackage{amsmath,amsfonts}
\usepackage[linesnumbered]{algorithm2e}
\usepackage{array}
\usepackage[caption=false,font=normalsize,labelfont=sf,textfont=sf]{subfig}
\usepackage{textcomp}
\usepackage{stfloats}
\usepackage{url}
\usepackage{verbatim}
\usepackage{graphicx}
\usepackage{cite}
\usepackage{float}  
\usepackage{enumitem} 
\usepackage{graphicx}  
\usepackage{tabularx}
\usepackage{array}
\usepackage{makecell} 
\usepackage{caption} 

\begin{document}

\title{Overcoming Challenges of Partial Client Participation in Federated Learning : A Comprehensive Review}

\author{Mrinmay Sen, Shruti Aparna, Rohit Agarwal, Chalavadi Krishna Mohan,~\IEEEmembership{Senior Member,~IEEE}
\thanks{Mrinmay Sen is with the Department of Artificial Intelligence, Indian Institute of Technology Hyderabad, India and the Department of Computing Technologies, Swinburne University of Technology, Hawthorn, australia (e-mail: ai20resch11001@iith.ac.in).}
\thanks{Shruti Aparna is with the Department of Physics, Indian Institute of Technology Delhi, India.}
\thanks{Rohit Agarwal  is with the Department of Energy Science and Engineering, Indian Institute of Technology Delhi, India.}
\thanks{C Krishna Mohan is with the Department of Computer Science and Engineering, Indian Institute of Technology Hyderabad, India.}

\thanks{Manuscript received April 19, 2021; revised August 16, 2021.}}

\markboth{Journal of \LaTeX\ Class Files,~Vol.~14, No.~8, August~2021}%
{Shell \MakeLowercase{\textit{et al.}}: A Sample Article Using IEEEtran.cls for IEEE Journals}

\maketitle

\begin{abstract}
Federated Learning (FL) is a learning mechanism that falls under the distributed training umbrella, which collaboratively trains a shared global model without disclosing the raw data from different clients. This paper presents an extensive survey on the impact of partial client participation in federated learning. While much of the existing research focuses on addressing issues such as generalization, robustness, and fairness caused by data heterogeneity under the assumption of full client participation, limited attention has been given to the practical and theoretical challenges arising from partial client participation, which is common in real-world scenarios. This survey provides an in-depth review of existing FL methods designed to cope with partial client participation. We offer a comprehensive analysis supported by theoretical insights and empirical findings, along with a structured categorization of these methods, highlighting their respective advantages and disadvantages.

\end{abstract}


\section{Introduction}
\IEEEPARstart{T}{he} ongoing advancement of machine learning has caused significant changes across several study domains and produced profound effects on all aspects of societal and industrial sectors, encompassing computer vision, natural language processing, multimodal learning, medical analysis, and beyond~\cite{liu2024systematic,wen2023survey}. These developments have resulted in significant advancements in fields including image recognition, language translation, speech synthesis, and disease diagnosis. The exceptional performance of deep learning models is significantly dependent on the accessibility of extensive, high-quality datasets~\cite{woisetschlaeger2024,li2020generalizedfedavg}. The training of robust deep learning models often involves centralized data aggregation, raising increasing concerns about data privacy and legislative limitations in contemporary culture~\cite{li2019survey}.

In response to these concerns, governments and organizations have implemented many data protection laws, such as the Information Technology (Reasonable Security Practices and Procedures and Sensitive Personal Data or Information) Rules, 2011 in India, General Data Protection Regulation (GDPR) in Europe and the Health Insurance Portability and Accountability Act (HIPAA) in the United States. These restrictions regulate the gathering, dissemination, and use of sensitive personal data, complicating the application of conventional centralised training paradigms in contexts where data is intrinsically private and dispersed among several entities~\cite{li2019survey,wen2023survey}. Thus, the necessity for privacy-preserving learning paradigms has become essential~\cite{liu2024systematic,li2020generalizedfedavg}.

\subsection{The Rise of Federated Learning}
Federated Learning (FL) has become a prominent solution to the challenges of centralized learning, since it enables the training of a global model for various participants without the need to centralize data owned by distributed entities~\cite{li2019survey,li2020generalizedfedavg,liu2024systematic}. In the traditional federated learning framework, data remains on the clients' local devices and only model updates, such as gradients or weights, are sent to a central server. The server combines updates from several clients to formulate a global model, which is then redistributed back to the clients for additional optimization~\cite{mcmahan2017fedavg,woisetschlaeger2024}. Federated learning comprises two fundamental processes: 

\textbf{Server-Side Collaboration:} A central server gathers locally trained model updates from clients, aggregates them (often employing techniques such as Federated Averaging (FedAvg)), and redistributes the revised global model~\cite{mcmahan2017fedavg,li2020generalizedfedavg}. 

\textbf{Client-Side Optimization:} Each client enhances the global model utilizing their local data. This procedure is executed independently on clients' confidential datasets~\cite{li2019survey,liu2024systematic}. 

Federated Learning (FL) has exhibited its potential in applications including edge computing, smart healthcare, autonomous driving, and Internet of Things (IoT) networks by facilitating collaborative learning without data centralization~\cite{wen2023survey,li2019survey,safari2023}.

\subsection{The Challenge of Partial Participation}
Federated learning offers a novel approach to distributed training, although it presents significant issues due to its dependence on the active involvement of clients. A fundamental assumption in traditional federated learning systems is that all clients engage in each communication round~\cite{mcmahan2017fedavg,li2019survey}. Nevertheless, in practical situations, this assumption is invalidated by other circumstances, such as \textit{Resource Heterogeneity}, where clients frequently exhibit substantial disparities in processing power, battery longevity, network bandwidth, and memory capacity~\cite{liu2024systematic,wen2023survey,li2020generalizedfedavg}. Devices like smartphones, IoT sensors, and edge servers may be deficient in resources necessary for model training or regular communication of updates~\cite{woisetschlaeger2024}. Unreliable or inconsistent network connections may hinder some clients from engaging in specific training sessions, especially in rural or low-bandwidth settings~\cite{li2019survey,liu2024systematic}. Clients may not consistently possess data readily accessible for training. This is prevalent in dynamic settings where data is produced progressively or where privacy issues limit its utilization~\cite{wen2023survey,li2019survey}.

As the client base in federated learning systems expands, simultaneous participation by all clients becomes unfeasible due to communication constraints and heightened aggregation expenses~\cite{fedvarp2022,li2021fedamd}. In specific federated systems, clients can voluntarily join or exit, resulting in inconsistent participation patterns~\cite{fedcm2022,woisetschlaeger2024}. These qualities result in the problem of restricted customer participation, whereby only a fraction of clients participates in each training round. The variation in client participation can profoundly influence the performance, efficiency, and convergence of federated learning models~\cite{liu2024systematic,wen2023survey}. Uneven client engagement, particularly in non-IID data distributions, may skew the global model in favor of more active users~\cite{fedvarp2022,mifa2023}. This diminishes the model's generalisability to the complete client population. Regular client dropout and inconsistent participation may result in unstable or delayed convergence, impairing model performance~\cite{karimireddy2020scaffold,li2020generalizedfedavg}.

\subsection{Motivation for This Survey}
Although partial client participation is becoming increasingly significant, the majority of current federated learning surveys predominantly address concerns such as generalization, robustness, and fairness, neglecting the unique issues stemming from heterogeneity in client participation~\cite{wen2023survey,li2020generalizedfedavg,liu2024systematic}.
This survey provides a thorough examination of methodologies, tactics, and frameworks aimed at managing partial participation in federated learning. Specifically, we aim to emphasis the distinct issues presented by limited client participation in federated learning systems. Deliver a thorough classification of current methodologies, encompassing client selection tactics, adaptive aggregation techniques, and communication-efficient solutions~\cite{fedvarp2022,li2021fedamd,karimireddy2020scaffold}.
Analyses the impact of partial participation on essential elements such as convergence stability, model generalization, and resource efficiency.
Identify current challenges and potential opportunities to further research on partial participation in federated learning.

\subsection{Contributions}
In this survey, we make the following contributions:
\begin{itemize}
    \item Systematic Categorization: We categories current methods for managing partial client participation into several areas, such as client selection, aggregation strategies, communication optimization, and resource-aware scheduling.
    \item Comparative Analysis: We do a comprehensive evaluation of current methodologies, including their advantages, drawbacks, and appropriateness for diverse FL contexts.
    
    \item Practical Insights: We examine real-world case studies and difficulties where partial participation is a prevalent issue, including mobile edge networks, healthcare federated systems, and IoT contexts.
    
    \item Benchmarking: We delineate benchmark datasets, assessment measures, and experimental frameworks employed to evaluate methodologies addressing limited participation.
    
    \item Future Directions: We emphasis open research problems and prospective opportunities to enhance the field of federated learning in the context of partial participation.
\end{itemize}

\section{Problem Definition}
Following the standard Federated Learning setup, suppose there is a set \( C \) of \( M \) clients (indexed by \( i \)), each with a private dataset \( \mathcal{D}_i \). Let \( N_i = |\mathcal{D}_i| \) denote the size of the private dataset for the \( i \)-th client. We denote the global model parameters by \( \mathbf{w} \).
Federated Learning aims to learn an optimal global model \( \mathbf{w}^* \) that minimizes the weighted empirical loss across all clients as shown in Eq. \ref{eq:1}.

\begin{equation}
    \label{eq:1}
    \textbf{w}^* = \arg\min_{\textbf{w}} \sum_{i=1}^M \alpha_i F_i(\textbf{w}, \mathcal{D}_i)
\end{equation}

where $\alpha_i$ denotes the pre-allocated aggregation weight \newline
($\sum_{i=1}^M \alpha_i = 1$) and is normally based on the data scale as $\alpha_i = \frac{N_i}{\sum_{j=1}^M N_j}$ and $F_i(\textbf{w}, \mathcal{D}_i)$ represents the empirical loss of $i$-th client.

In a federated learning (FL) system with $M$ total clients, we consider the challenge of \textbf{partial client participation}, where only a subset $\widetilde{C}$ of $m < M$ clients participates in each training round instead of all $M$ clients. This scenario arises due to factors such as resource constraints, network issues, or selective sampling processes~\cite{liu2024systematic,fedvarp2022,li2021fedamd}. In each communication round in FL, the objective function is reformulated as shown in Eq. \ref{eq:2}, where $\sum_{i=1}^{|\widetilde{C}|} \alpha_i = 1$ and $\alpha_i = \frac{N_i}{\sum_{j=1}^{|\widetilde{C}|} N_j}$. 

\begin{equation}
    \label{eq:2}
    \textbf{w}^* = \arg\min_{\textbf{w}} \sum_{i=1}^{|\widetilde{C}|} \alpha_i F_i(\textbf{w}, \mathcal{D}_i)
\end{equation}

\section{Approaches to Address Partial Participation in Federated Learning}
To address the challenges posed by \textbf{partial client participation}, various algorithms have been proposed in the literature. Below, we discuss the major methods and their representative works.

\subsection{\bf Embracing FL}
Embracing FL ~\cite{lee2024embracing} divides the model into layers and assigns different layers to clients based on their computational capacity. Unlike traditional FL, where each client is expected to train a complete local model, Embracing FL assigns specific portions of the model to be trained by different clients based on their resource capabilities. The framework primarily focuses two things. One is Output-Side Layer Training for Weak Clients, where weak clients train only the output-side layers of the model. This strategy leverages the insight that output-side layers tend to learn more critical and unique data representations, while input-side layers often generalize across clients. Another is Input-Side Layer Training for Strong Clients, where strong clients are tasked with training the entire model, ensuring that the foundational layers remain robust and effective. The operational workflow of Embracing FL consists of the following steps:
\begin{enumerate}[label= -]
    \item \textbf{Model Segmentation:} The global model is divided into input-side and output-side layers. Weak clients are assigned only the output-side layers.
    \item \textbf{Multi-Step Forward Pass:} Weak clients receive the input-side layers from the server, process them in a multi-step forward pass using their local data, and store the intermediate activations. These stored activations serve as the input for training the output-side layers.
    \item \textbf{Local Training on Output-Side Layers:} Weak clients perform gradient updates only on the output-side layers, significantly reducing memory and computational overhead.
    \item \textbf{Model Aggregation:} After local training, all clients upload their trained layers to the server, where they are aggregated into a unified global model.
\end{enumerate}
\subsubsection{Theoretical Analysis of Embracing FL} 
By training only specific layers, Embracing FL ensures:

\begin{itemize}[label=-]
    \item \textbf{Heterogeneity:} Accommodates clients with varying computational power.
    \item \textbf{Convergence:} Maintains training stability despite partial participation.
    \item \textbf{Resource Efficiency:} Reduces memory and computation requirements on weak clients.
\end{itemize}

\subsubsection{Experiment}
EmbracingFL was evaluated on CIFAR-10 (ResNet20), FEMNIST (CNN), and IMDB (LSTM) under non-IID settings and varying client capacities. Non-IID distributions were simulated using a Dirichlet distribution ($\alpha = 0.1$), and client heterogeneity was introduced by categorizing clients as strong, moderate, or weak based on model capacity. Despite extreme variations in data and device capabilities, Embracing FL achieved competitive accuracy across all benchmarks. Notably, it maintained performance even when weak clients used models with as little as $2\%$ of the full model capacity. These results highlight EmbracingFL’s robustness to both statistical and system heterogeneity in federated learning.

\subsubsection{Limitation}
Despite its efficiency, EmbracingFL introduces several challenges. It increases server-side complexity due to model segmentation and heterogeneous aggregation. Synchronization between strong and weak clients can lead to delays, and less frequent updates to input-side layers may hinder convergence. The use of intermediate activations raises privacy concerns, and the approach may not generalize well to complex architectures. Additionally, uneven client contributions can bias the global model toward strong clients' data.

\subsection{\bf Cyclic Client Participation (CyCP)} 

CyCp \cite{cho2023cycp} introduces an efficient client sampling strategy for FL that accounts for the natural cyclic availability patterns of clients in practical systems. Unlike traditional client sampling methods that uniformly or randomly select clients, CyCP divides clients into predefined groups that participate cyclically, reflecting real-world scenarios like time-zone-based availability or device charging patterns. The operational flow of FedAvg with CyCP can be described as follows:
\begin{enumerate} [label=-]
    \item \textbf{Client grouping}
    \begin{itemize}
        \item Divide the total clients $M$ into $K$ non-overlapping groups:
        \[\sigma^{(1)}, \sigma^{(2)}, \dots, \sigma^{(K)},\]
        where each group contains $\frac{M}{K}$ clients.
        \item Predetermine a cyclic order in which these groups will become available for training.
    \end{itemize}
    \item \textbf{Communication Rounds}
    \begin{itemize}
        \item Each training cycle consists of K communication rounds, covering all groups once. The index k represents the cycle, and i represents the current group in the cycle.
    \end{itemize}
    \item \textbf{Server and Client Interaction}
    \begin{itemize}
        \item \textbf{Model Distribution:} The server sends the current global model $w^{(k, i-1)}$ to N randomly sampled clients from the active group $\sigma(i)$
        \item \textbf{Local Training:} Each selected client initializes its local model with the global model $w^{(k, i-1)}$ received from the server. The clients then perform local updates using one of three strategies based on their computational capabilities. In Gradient Descent (GD), clients compute the full gradient and perform a single update. With Stochastic Gradient Descent (SGD), clients use mini-batch stochastic updates over $\tau$ local steps to refine the model. In Shuffled SGD (SSGD), clients partition their local data into disjoint subsets, shuffle them, and sequentially perform updates for better convergence. Once the local updates are complete, they are aggregated into a combined update, $\Delta^{(k, i-1)}$ which is sent back to the server for global model refinement.
        \item \textbf{Model Aggregation: }The server computes the new global model as $w^{(k, i)}$ = $w^{(k, i-1)}$ + $\Delta^{(k, i-1)}$
        
    \end{itemize}
    \item \textbf{Cyclic Progression: }After all K rounds in a cycle, the server starts the next cycle with updated models and iterates until convergence.
\end{enumerate}
\subsubsection{Theoretical Analysis of CyCP}
This method employs a cyclic schedule where clients are selected in a structured manner rather than randomly. Key features include:
    \begin{itemize}[label=-]
        \item \textbf{Heterogeneity:} Provides a systematic way to include diverse clients.
        \item \textbf{Convergence:} Achieves faster convergence by balancing participation across rounds.
        \item \textbf{Benchmark:} Demonstrates effectiveness on standard FL datasets.
    \end{itemize}

\subsubsection{Experiment}
We evaluate the performance of Cyclic Client Participation (CyCP) in FedAvg using FMNIST and EMNIST datasets under varying data heterogeneity and values of K, the number of client groups per cycle. Data was partitioned using a Dirichlet distribution ($\alpha$) to simulate non-IID settings. Experiments were conducted with MLPs and multiple local update procedures (GD, SGD, SSGD), with grid search over learning rate, batch size, and local iterations for optimal performance. Results demonstrate that increasing K consistently improves test accuracy across all client procedures. Under high heterogeneity ($\alpha$=0.5), CyCP with higher K achieves up to 5–10\% better accuracy than standard FedAvg (K=1). With lower heterogeneity ($\alpha$=2.0), the gain reduces to 2–8\%, aligning with theoretical expectations. However, large K can introduce accuracy oscillations due to increased inter-group heterogeneity during training.

\subsubsection{Limitation}
While CyCP ensures structured and fair client involvement, it also introduces certain limitations. The fixed grouping of clients may not adapt well to dynamic availability or client dropouts, leading to missed updates or reduced diversity in training. Additionally, clients in the same group may share similar data distributions (e.g., due to geographical or temporal alignment), which can increase intra-group data homogeneity and slow global convergence. The cyclic scheduling may also result in longer intervals between participation for certain clients, delaying their influence on the global model and making adaptation to non-stationary data more difficult.


\subsection{\bf Fed-EF}
In federated learning, communication cost is a significant bottleneck, particularly when models involve millions of parameters. Biased gradient compression can reduce this cost but often introduces non-vanishing errors that hinder convergence. The Fed-EF \cite{pauloski2020fedef} framework incorporates error feedback (EF) to address this issue. It includes two variants: Fed-EF-SGD (with stochastic gradient descent) and Fed-EF-AMS (using AMSGrad, an adaptive optimizer). The framework ensures convergence rates comparable to full-precision FL while substantially reducing communication overhead. The operational steps in the Fed-EF framework are as follows:
\begin{enumerate}[label = -]
    \item The central server initializes the global model parameters ($\theta_1$) and sets the error accumulators ($e_{1,i}$) for all clients. Additional parameters, such as momentum ($m_0$) and variance trackers ($v_0$,$\hat{v}_0$), are initialized if the AMSGrad variant is used.
    \item At the beginning of each communication round t, the server broadcasts the current global model ($\theta_t$) to all participating clients.
    \item Each participating client uses its local dataset to perform K local training steps. During each step, the client updates its local model ($\theta_{(t,i)}$) using stochastic gradient descent (SGD), with the learning rate $\eta_l$.
    \item After completing the K local training steps, the client computes its local model update $\Delta_{(t,i)}$ = $\theta_{(t,i)}^{(k+1)}$ - $\theta_t$).
    \item To reduce communication overhead, the client applies a biased compression operator ($\mathcal{C}$) to the sum of the local update ($\Delta_{t,i}$) and its error accumulator ($e_{t,i}$). The compressed update ($\widetilde{\Delta}_{t,i}$) is then obtained.
    \item The client updates its error accumulator ($e_{t+1,i}$) to account for the compression-induced error using the formula:
    \[
    e_{t+1,i} = e_{t,i} + \Delta_{t,i} - \widetilde{\Delta}_{t,i}.
    \]
    \item The client sends the compressed update ($\widetilde{\Delta}_{t,i}$) to the server.
        \item Upon receiving the compressed updates from all participating clients, the server aggregates them to compute the global update:
    \[
    \widetilde{\Delta}_t = \frac{1}{n} \sum_{i=1}^{n} \widetilde{\Delta}_{t,i},
    \]
    where $n$ is the number of active clients.
    \item The server updates the global model parameters using the aggregated update. For the SGD variant (Fed-EF-SGD), the global model is updated as:
    \[
    \theta_{t+1} = \theta_t - \eta \widetilde{\Delta}_t.
    \]
    For the AMSGrad variant (Fed-EF-AMS), the server uses adaptive updates:
    \begin{itemize}
        \item $m_t = \beta_1 m_{t-1} + (1 - \beta_1) \widetilde{\Delta}_t,$
        \item $v_t = \beta_2 v_{t-1} + (1 - \beta_2) (\widetilde{\Delta}_t)^2,$
        \item $\hat{v}_t = \max(v_t, \hat{v}_{t-1}),$
        \item $\theta_{t+1} = \theta_t - \frac{m_t}{\sqrt{\hat{v}_t} + \epsilon}.$
    \end{itemize}
    \item The above steps are repeated for the $T$ communication rounds until the global model converges to a stationary point or meets the desired accuracy. 
\end{enumerate}
\subsubsection{Theoretical Analysis of Fed-EF}
Fed-EF introduces error feedback mechanisms to mitigate the bias caused by compressed gradients. This method focuses on:
    \begin{itemize}[label=-]
        \item \textbf{Convergence:} Ensures stable training under compressed updates.
        \item \textbf{Resource Efficiency:} Reduces communication costs, enabling lightweight participation for clients with limited bandwidth.
    \end{itemize}
    
\subsubsection{Experiment}
The evaluation of Fed-EF and its variants is carried out on three standard federated learning datasets: MNIST, FMNIST, and CIFAR-10. Both MNIST and FMNIST consist of 28×28 grayscale images, while CIFAR-10 includes 32×32 RGB images across 10 object categories. For CIFAR-10, standard preprocessing techniques such as random cropping, horizontal flipping, and pixel normalization are applied. A federated setting with 200 clients is simulated, where data is made highly non-IID by assigning each client two randomly selected shards, each containing samples from a single class. Each federated round involves partial participation with a sampling ratio p=m/n, where m clients are selected uniformly at random. Each selected client performs one local epoch with a mini-batch size of 32, amounting to 10 local updates per round. Fed-EF variants are tested with Sign and Top-k compressors, using \( k \in \{0.001, 0.01, 0.05\} \), and an additional heavy-Sign method, which applies Sign after Top-k, further increasing compression \( k \in \{0.01, 0.05, 0.1\} \). Comparisons are made against baseline methods using full-precision gradients and stochastic quantization (“Stoc”) without error feedback for bit-widths \( b \in \{1,2,4\} \). Communication efficiency is reported as the average number of bits transmitted per client, assuming 32-bit precision. All metrics are averaged over multiple independent runs for reliability.\\

\subsubsection{Limitations}
Despite reducing communication overhead, Fed-EF has limitations. The use of biased compression requires careful tuning of error feedback mechanisms to avoid instability. Accumulated errors may amplify if updates are noisy or poorly estimated, especially in non-iid settings. Additionally, maintaining per-client error accumulators increases local memory usage, which may burden resource-constrained devices. Lastly, adaptive variants like Fed-EF-AMS introduce extra complexity in hyperparameter tuning and may be sensitive to learning rate schedules.



\subsection{\bf GradMA} 
The GradMA \cite{gradma2023} approach for handling partial participation in federated learning is designed to alleviate catastrophic forgetting while optimizing the server and worker updates. The method combines continual learning techniques with federated learning to improve convergence and accuracy in scenarios with data heterogeneity and random worker participation. Below is an outline of the detailed operational steps.
\begin{enumerate}[label =-]
    \item Initialization includes assigning initial states to the server and workers, specifying parameters like learning rates and memory sizes, and preparing data structures for memory management.
    \item Server-side operations involve sampling active workers in each communication round, transmitting the current global model to these workers, and maintaining a memory state that records accumulated updates from all workers to account for contributions from inactive workers.
    \item Worker-side operations include performing local training using quadratic programming (QP) to adaptively adjust gradients based on previous gradients, global model information, and local parameters, ensuring updates align with local and global objectives, and sending updated model differences back to the server.
    \item The update process on the server refines the global model using received updates, applies QP to determine the update direction based on the memory state that incorporates updates from both active and inactive workers, and employs a memory reduction strategy to efficiently manage storage when the memory buffer reaches capacity. 
    \item The memory reduction strategy controls the memory size by discarding updates from less significant workers, ensuring efficiency while maintaining performance.
\end{enumerate}
\subsubsection{Theoretical Analysis of GradMA} 
This method maintains a gradient memory to store updates from previous rounds, ensuring training progress even when clients drop out. Key advantages include:
    \begin{itemize}[label=-]
        \item \textbf{Heterogeneity:} Handles varying participation effectively.
        \item \textbf{Resource Efficiency:} Reduces redundant computations by leveraging stored gradients.
        \item \textbf{Benchmark:} Demonstrates performance on benchmark datasets.
    \end{itemize}
    
\subsubsection{Experiment}
To evaluate the GradMA framework, experiments are conducted under a centralized federated learning setup involving 100 workers. The method is assessed through ablation studies by isolating the Worker Update and Server Update components using two algorithmic variants: GradMA-W and GradMA-S. Comparisons are made with a wide range of baselines. These baselines are grouped based on whether their modifications target the worker side, the server side, or both. The experiments consider different levels of partial participation with the number of sampled clients per round \( S \in \{5,10,50\} \), and varying degrees of data heterogeneity using Dirichlet distribution with concentration parameter \( w \in \{0.01,0.1,1.0\} \). For the models, MNIST uses a neural network with three fully connected layers, CIFAR-10 uses LeNet-5, CIFAR-100 uses VGG-11, and TinyImageNet employs ResNet-20. All methods are evaluated in terms of test accuracy and communication efficiency, with synchronization interval fixed at I=5. The impact of control parameters \( (\beta_1, \beta_2) \) and memory size m is further studied empirically on top of the GradMA-S variant.
\subsubsection{Limitation}
GradMA introduces computational overhead due to solving quadratic programming (QP) problems on both the client and server sides, which may limit its scalability in resource-constrained environments. Maintaining gradient memory also increases storage demands, and the memory reduction strategy may risk discarding potentially useful updates. Additionally, the effectiveness of gradient correction relies heavily on accurate QP solutions, making the method sensitive to optimization inaccuracies and hyperparameter tuning.


\subsection{\bf FedAMD} FedAMD \cite{li2021fedamd} introduces anchor sampling, which divides the participating clients into two distinct groups: the anchor group and the miner group. In anchor group clients compute gradients using a large batch of local data. This helps estimate the optimal direction (or "bullseye") for global convergence. These gradients are then cached to guide subsequent updates. In miner group clients perform multiple small-batch local updates based on the global model's target direction. This approach accelerates the training process and ensures the global model improves iteratively by aggregating miner updates. Below is an outline of the detailed operational steps:
    \begin{enumerate}[label = -]
        \item \textbf{Client Selection:} A subset of clients is sampled for each training round. The selection probability is dynamically determined, distinguishing between anchor and miner roles.
        \item \textbf{Anchor Operations:} Anchors compute gradients with large batches to align with global convergence objectives.
        \item \textbf{Miner Operations:} Miners execute multiple small-batch local updates and submit the resulting model changes to the server.
        \item \textbf{Global Update:} The server integrates updates from miners and leverages cached gradients from anchors to refine the global model.
    \end{enumerate}
\subsubsection{Theoretical Analysis of FedAMD}
This method utilizes an anchor-based sampling strategy to select clients with diverse and representative updates. This approach improves:
    \begin{itemize}[label=-]
        \item \textbf{Heterogeneity:} Balances the impact of non-IID data.
        \item \textbf{Convergence:} Stabilizes learning by ensuring diverse client participation.
    \end{itemize}
    
\subsubsection{Experiment}
FedAMD is evaluated on two benchmark datasets: Fashion-MNIST and EMNIST-digits. A convolutional neural network (LeNet-5) is used for Fashion-MNIST, while a 2-layer multilayer perceptron (MLP) is applied to EMNIST. The experiments are conducted with 100 clients, and to simulate non-i.i.d. data distribution, each client is assigned samples from only two distinct classes. The default configuration uses K=10, local batch size \( b' \), and local dataset size b. Unless otherwise stated, hyperparameters such as learning rates are optimized to achieve the best performance. All reported results represent the average over three runs with different random seeds. In addition to comparisons with relevant baseline methods, the impact of the participation probability sequence \( \{p_t\}_{t \geq 0} \) is examined, while further numerical analysis involving the number of local updates is provided in the appendix.

\subsubsection{Limitation}
While FedAMD improves convergence and stability through its anchor-miner client structure, it introduces several limitations. First, the need to compute large-batch gradients for anchor clients increases computational and communication overhead, which may be impractical for low-resource devices. Additionally, the accuracy and effectiveness of anchor updates rely heavily on representative data sampling, which may not always be feasible in non-IID settings. The complexity of coordinating two client roles and managing cached gradients also increases implementation difficulty. Finally, if anchor selection is suboptimal or imbalanced, the guidance provided to miners may degrade model performance.\\

\subsection{\bf Generalized FedAvg}
The generalized Federated Averaging (FedAvg) \cite{li2020generalizedfedavg} with two-sided learning rates demonstrates that it is possible to achieve a linear speedup in convergence even under non-IID (non-independent, non-identically distributed) data and partial worker participation scenarios. The algorithm allows a subset of workers to participate in each communication round instead of requiring full participation. It employs random sampling to select a subset of workers at each round, ensuring a balance between computational efficiency and maintaining convergence properties. The method adjusts learning rates dynamically to mitigate the impact of heterogeneity in the data and system. Below is an outline of the detailed operational steps:
    \begin{enumerate}[label=-]
        \item \textbf{Worker Sampling:} At each communication round, a subset of workers is randomly selected. The selection probabilities ensure that the sampled workers represent the overall population adequately.
        \item \textbf{Local Updates:} Selected workers perform multiple local stochastic gradient descent (SGD) steps to minimize their local loss functions using their individual datasets.
        \item \textbf{Parameter Aggregation:} Each selected worker sends an accumulated parameter difference to the server. The server aggregates these updates.
        \item \textbf{Global Update:} The global model is updated by the server using the aggregated updates, with learning rates tuned to handle variance caused by partial participation and data heterogeneity.
        \item \textbf{Broadcast Model:} The updated global model is broadcast back to the workers.
    \end{enumerate}
    This framework ensures convergence at a rate of $O\left(\frac{\sqrt{K}}{\sqrt{nT}} + \frac{1}{T}\right)$ under partial worker participation, where K is the number of local steps, n is the number of participating workers per round, and T is the total number of communication rounds.\\
\subsubsection{Theoretical Analysis of Generalized FedAvg}
It incorporates an adaptive learning rate mechanism to improve the aggregation process. It focuses on:
    \begin{itemize}[label=-]
        \item \textbf{Convergence:} Speeds up training by tuning learning rates based on participation levels.
        \item \textbf{Resource Efficiency:} Reduces unnecessary updates from weaker clients.
    \end{itemize}

\subsubsection{Experiment}
To evaluate the generalized FedAvg algorithm, experiments are conducted on both non-i.i.d. and i.i.d. versions of the MNIST and CIFAR-10 datasets. Three model architectures are used: logistic regression (LR), a fully connected neural network with two hidden layers (2NN), and a convolutional neural network (CNN). For CIFAR-10, a ResNet architecture is used. The non-i.i.d. data distributions are generated by partitioning MNIST such that each of the 100 workers receives samples associated with a limited number of digit classes (e.g., 1, 2, 5, or 10 digits per worker), controlling the degree of heterogeneity. Experiments are also conducted under both full and partial worker participation, with subsets of workers (e.g., 10, 50, or 100) participating in each communication round. Additional evaluations are performed to assess the impact of varying local training steps. Results demonstrate that the convergence speed and test accuracy are significantly influenced by the degree of data heterogeneity, number of participating workers, and number of local training epochs. Further experimental results and comparisons are provided in the supplementary material.
\subsubsection{Limitation}
Despite its enhancements over the classical FedAvg, the Generalized FedAvg algorithm presents certain limitations. The reliance on carefully tuned two-sided learning rates introduces sensitivity in performance—suboptimal choices can hinder convergence or even destabilize training. Additionally, the random worker sampling strategy, while efficient, may occasionally lead to under representation of crucial client data, especially in highly skewed non-IID settings. Since the method assumes statistical adequacy of the sampled clients over rounds, it may perform poorly if the sampling process is biased or sparse. Furthermore, the aggregation of accumulated parameter differences rather than full models can amplify variance in updates, particularly when local steps (K) are large. This can lead to slower convergence or oscillatory behavior in heterogeneous systems.

\subsection{\bf SAFARI or SA-FL}

To address challenges posed by incomplete client participation, SA-FL \cite{safari2023} introduces an auxiliary dataset maintained at the server. This dataset is designed to mimic the distribution of data from non-participating clients, thereby mitigating deviations caused by data heterogeneity. This approach ensures that the system remains PAC-learnable (Probably Approximately Correct learnable) under such participation constraints. Below is an outline of the detailed operational steps:
\begin{enumerate}[label = -]
     \item Server Setup: The server maintains an auxiliary dataset representing the overall data distribution, compensating for absent client data.
    \item Client Sampling: A subset of clients is selected for participation in each training round, similar to traditional federated learning.
    \item Local Updates: Participating clients perform local training and share model updates with the server.
    \item Server Updates: The server concurrently trains on its auxiliary dataset and aggregates updates received from clients.
    \item Global Model Update: The server integrates client updates with its own to refine the global model.
\end{enumerate}
This hybrid approach improves model performance and guarantees convergence rates even under system heterogeneity, with comparable communication and privacy efficiency to traditional federated learning methods.
\subsubsection{Theoretical Analysis of SAFARI}
A server-assisted federated learning framework that provides additional computational support to resource-constrained clients. Key contributions include:
    \begin{itemize}[label=-]
        \item \textbf{Heterogeneity:} Handles variations in client capabilities through server assistance.
        \item \textbf{Convergence:} Improves stability by reducing client dropout impact.
        \item \textbf{Resource Efficiency:} Reduces the burden on clients through server interventions.
        \item \textbf{Benchmark:} Validated on benchmark datasets to demonstrate effectiveness.
    \end{itemize}
    
\subsubsection{Experiment}
In evaluating the SAFARI algorithm, a comprehensive experimental setup is designed that closely mirrors practical federated learning (FL) environments characterized by data heterogeneity and partial client participation. Two standard benchmark datasets are employed: MNIST, with a multinomial logistic regression model, and CIFAR-10, with a convolutional neural network. The system consists of 10 clients, of which 5 are randomly selected to participate in each communication round. To simulate statistical heterogeneity, each client’s local dataset is restricted to samples from only p out of the 10 total classes; smaller values of p (e.g., 1 or 2) correspond to more heterogeneous, non-i.i.d. distributions. System heterogeneity is introduced through an incomplete participation index s, representing the number of clients excluded from the training pool. In addition, the server is equipped with a small auxiliary dataset drawn i.i.d. from the global distribution, with sizes ranging from 50 to 1000 samples for MNIST and up to 5000 samples for CIFAR-10. SAFARI is evaluated against two baselines: the standard FedAvg algorithm and centralized SGD trained solely on the server’s auxiliary data. Test accuracy across communication rounds is used as the primary performance metric. This setup highlights SAFARI’s ability to maintain strong generalization performance even in the presence of highly skewed data distributions and limited client participation.
\subsubsection{Limitation}
While SAFARI demonstrates notable improvements over conventional FL approaches under incomplete client participation, it does have certain limitations. Most notably, the approach relies on the availability of an auxiliary dataset at the server that is representative of the global data distribution. Although the required dataset size is relatively small, acquiring even a modest number of labeled samples that align well with non-participating clients' data may not always be feasible in privacy-sensitive or domain-specific applications. Additionally, SAFARI assumes that the auxiliary data distribution is i.i.d. with the global distribution, which may not hold in practical scenarios, potentially degrading its performance. From a theoretical perspective, although SAFARI provides PAC-learnability guarantees, the bounds rely on conditions such as  \( (\alpha, \beta) \) -positive relatedness between distributions, which can be difficult to verify or satisfy in real-world deployments. Moreover, the algorithm introduces new hyperparameters (e.g., $c_t$) for rescaling client updates) and added computation on the server side, which may require careful tuning and could increase system complexity. These factors highlight the trade-off between robustness and practicality in deploying SAFARI in diverse federated settings.

\subsection{\bf Memory-augmented Impatient Federated Averaging (MIFA)} 
Memory-augmented methods use auxiliary memory to compensate for incomplete participation and stabilize the training process. MIFA \cite{mifa2023} operates by maintaining a memory of the latest updates from all devices, active or inactive. When devices are unavailable, their most recent updates stored in memory are used as surrogates. This avoids delays caused by waiting for responses and ensures progress even with missing updates. The algorithm is robust to non-stationary and adversarial unavailability patterns, making it practical for real-world federated learning setups. Below is an outline of the detailed operational steps:
\begin{enumerate}[label =-]
    \item Model Broadcast: At the beginning of each round, the server broadcasts the latest global model to all devices.
    \item Local Computation: Active devices perform several local updates using their private data and return their updated gradients to the server.
    \item Memory Update: For each active device, the server stores the received update in memory. For inactive devices, their previously stored updates are retained.
    \item Global Model Update: The server aggregates updates from all devices—using current updates from active devices and memorized updates from inactive ones—to refine the global model.
\end{enumerate}
This approach eliminates excessive latency caused by inactive devices while correcting the gradient bias that might arise from using stale updates. Theoretical analysis demonstrates that MIFA achieves minimax optimal convergence rates for both convex and non-convex objectives.

\subsubsection{Theoretical Analysis}
This method employs memory augmentation to store model updates and ensure smoother training despite client variability. It addresses:
    \begin{itemize}[label=-]
        \item \textbf{Heterogeneity:} Supports diverse client participation patterns.
        \item \textbf{Convergence:} Improves training stability under irregular participation.
    \end{itemize}
\subsubsection{Experiment}
The experimental setup for evaluating MIFA involved training models on two standard computer vision datasets, MNIST and CIFAR-10, using a federated learning framework with 100 devices. To simulate a highly non-i.i.d. environment, each device was assigned data from only two classes. Device availability followed an i.i.d. Bernoulli model, where each device had a different probability of participation, with lower probabilities assigned to devices holding data from smaller class labels, increasing the heterogeneity of participation. A logistic regression model was used for MNIST and a LeNet-5 convolutional neural network for CIFAR-10. Local training involved two epochs per communication round with a batch size of 100. The learning rate decayed over time, and weight decay was applied for regularization. All experiments were run using four NVIDIA RTX 2080 Ti GPUs and repeated over five random seeds to ensure statistical reliability. MIFA’s performance was compared against biased FedAvg, FedAvg with importance sampling, and FedAvg with device sampling, showing superior convergence speed and robustness to device unavailability.
\subsubsection{Limitation}
While MIFA demonstrates strong performance in handling arbitrary device unavailability and achieves minimax optimal convergence rates, several limitations remain. First, the algorithm assumes that each device can store its most recent update, which may not be feasible for extremely resource-constrained devices due to memory or storage limitations, especially for large models. Second, while MIFA is theoretically robust to adversarial or non-stationary availability patterns, its performance in highly dynamic real-world network conditions with bursty availability and variable communication bandwidths has not been thoroughly explored. Additionally, the convergence analysis for non-convex objectives relies on strong assumptions, such as bounded gradient noise and gradient dissimilarity, which may not hold in practical scenarios. Lastly, although the experimental results validate MIFA’s efficiency, they are limited to computer vision tasks; evaluating the method on a broader range of applications, such as natural language processing or time-series data, would help generalize its effectiveness.

\subsection{\bf Momentum-Based Optimization: FedCM}
FedCM \cite{fedcm2022} improves convergence and stability in federated learning by incorporating historical gradient information into the optimization process. These method effectively addresses challenges arising from partial participation and heterogeneity in FL systems. FedCM’s operational workflow includes:
\begin{enumerate}[label=-]
    \item \textbf{Server Momentum Update:} Maintains a global momentum term that integrates past gradients, providing a shared optimization direction.
    \item \textbf{Client Momentum Integration:} Clients update their local models by combining local gradients with the global momentum.
    \item \textbf{Model Aggregation:} Updated models are aggregated at the server, which also updates the global momentum term.
\end{enumerate}

\subsubsection{Theoretical Analysis}
FedCM introduces a global momentum term maintained at the server, which aggregates gradient information over multiple rounds. Key features include:
\begin{itemize}[label =-]
    \item \textbf{Strongly Convex Objectives}: FedCM achieves faster convergence rates than FedAvg by aligning local updates with global objectives.
    \item \textbf{Non-Convex Settings}: The momentum mechanism reduces oscillations in optimization paths, improving stability and convergence.
    \item \textbf{Participation Robustness}: Even with partial participation, the aggregated global momentum ensures consistent optimization direction.
\end{itemize}

\subsubsection{Experiment}
The performance of FedCM was evaluated on the CIFAR-10 and CIFAR-100 datasets under two federated learning scenarios: (I) 100 clients with 10\% participation, and (II) 500 clients with 2\% participation. In both settings, clients were activated independently with fixed probabilities (0.1 and 0.02, respectively). The datasets were partitioned in both i.i.d. and non-i.i.d. fashions, with the non-i.i.d. splits generated using a Dirichlet distribution with concentration parameter 0.6. Each client held a balanced number of data samples—500 in setting I and 100 in setting II. A standard ResNet-18 architecture with group normalization replaced batch normalization for improved training stability on distributed clients. The experiments compared FedCM against FedAvg, FedAdam, SCAFFOLD, and FedDyn over 4000 communication rounds. Results were reported in terms of test accuracy to assess convergence and generalization under varying degrees of client heterogeneity and participation rates.

\subsubsection{Limitation}

While FedCM offers notable improvements in handling client heterogeneity and low participation rates, it has several limitations. First, it introduces an additional communication overhead from the server to the clients due to the transmission of the global momentum term, which may become significant in bandwidth-constrained environments. Although this is mitigated by leveraging typically faster downlink speeds, it still adds complexity. Second, the effectiveness of FedCM relies heavily on the proper tuning of the hyperparameter $\alpha$, which balances local and global gradient information; poor tuning can lead to slower convergence or instability. Moreover, the method assumes synchronized communication rounds and uniform client availability probabilities, which may not reflect real-world, highly asynchronous federated settings. Lastly, while the experiments demonstrate strong results on image classification tasks, the generalizability of FedCM to other domains such as natural language processing or time series data remains to be thoroughly explored.

\subsection{\bf Variance Reduction for Partial Participation (FedVARP)}
FedVARP \cite{fedvarp2022} addresses the critical issue of variance caused by partial client participation in federated learning. This novel algorithm introduces server-side variance reduction techniques, enabling robust and efficient model training even when only a subset of clients participates in each round. The operational workflow of FedVarp is as follows:
\begin{enumerate}[label=-]
    \item \textbf{Setup Phase} The server starts by initializing the global model. It also creates a memory to store the most recent update from each client.
    \item \textbf{During Each Round} 
    \begin{itemize}
        \item Client Selection: The server starts by initializing the global model.
        \item Local Training at Clients: Selected clients download the current global model. They train locally on their private data using standard SGD for a few steps. Each client sends its local update back to the server.
        \item Aggregration at the Server: The server combines the updates received from the participating clients. To account for non-participating clients, the server uses the most recent stored updates from them. It blends current and past updates to form a new update for the global model, reducing variance caused by missing clients.
        \item Memory Update at the Server: For clients that participated, their latest updates replace the old ones in memory. For those that did not participate, the server keeps their previous updates unchanged.
    \end{itemize}
    \item \textbf{Key Features} Clients do not need to perform any extra computation or send additional information. The method reduces the training error caused by only a few clients participating each round. A memory-efficient version called ClusterFedVARP stores updates for groups of similar clients instead of individual ones, saving space on the server.

\end{enumerate}

\subsubsection{Theoretical Analysis}
FedVARP provides strong theoretical guarantees:
\begin{itemize}[label =-]
    \item \textbf{Convergence Rates}: FedVARP achieves faster convergence compared to FedAvg, with reduced variance ensuring stable optimization.
    \item \textbf{Scalability}: The clustering mechanism in ClusterFedVARP ensures scalability to large federations without significant memory overhead.
\end{itemize}
\subsubsection{Experiment}
The experimental evaluation of FedVARP was conducted on three federated learning tasks: image classification using the CIFAR-10 dataset with LeNet-5 and ResNet-18 architectures, and next-character prediction on the Shakespeare dataset using a single-layer GRU model. To simulate partial participation, only 5 clients were randomly selected in each round without replacement, resulting in a participation rate of 2\% for CIFAR-10 (250 clients) and less than 1\% for Shakespeare (over 1000 clients). Each client performed 5 local epochs per round using a batch size of 64. The server learning rate was fixed to 1, while the client learning rate was tuned over a logarithmic grid. For CIFAR-10 experiments, the data was non-i.i.d., with each client assigned data corresponding to 1 or 2 classes using a label-based shard strategy. In the Shakespeare dataset, each client represented a character from a play, and for ClusterFedVARP, clients were grouped by play to reduce memory requirements. All models were trained over several communication rounds and compared against FedAvg, SCAFFOLD, and MIFA in terms of convergence speed and test accuracy.

\subsubsection{Limitation}
While FedVARP effectively reduces the variance caused by partial client participation and achieves superior convergence without modifying client behavior, it has some notable limitations. The algorithm requires the server to store the latest update from every client, resulting in a memory overhead of order O(Nd), where N is the number of clients and d is the model dimension. This can become impractical in large-scale federated settings with thousands of clients or high-dimensional models. Although ClusterFedVARP addresses this issue through clustering, it introduces approximation error due to intra-cluster heterogeneity. Moreover, FedVARP assumes uniform random client sampling and consistent local update quality, which may not hold in real-world scenarios with highly dynamic client availability, system heterogeneity, or skewed data distributions. Finally, the algorithm’s performance has primarily been validated on vision and language modeling tasks; its effectiveness across other domains and under different types of system constraints remains to be explored.

\subsection{\bf Proximal Optimization for Partial Participation (FedProx)}
FedProx \cite{fedprox2018} addresses challenges associated with client heterogeneity and partial participation by introducing a proximal term to the local objective functions of clients, ensuring stability and scalability in federated learning systems. THe operational workflow of FedProx is given as follows:
\begin{enumerate} [label = -]
    \item \textbf{Intialization} The central server initializes the global model.A proximal coefficient \( \mu \) is set to control the influence of the proximal term.
    \item \textbf{Client Selection (Each Round)} A random subset of clients is selected for participation in the current communication round. The current global model is sent to all selected clients.
    \item \textbf{Local Computation (Client-Side)} Each selected client solves a proximal local objective, which is its own empirical loss function with an added quadratic regularization (proximal) term. Clients may perform variable amounts of local work, depending on their available resources (e.g., fewer epochs if resource-constrained).The result is an inexact minimizer, reflecting tolerance to system heterogeneity.
    \item \textbf{Communication} Each client sends back its local model update (after solving the proximal objective) to the server.
    \item \textbf{Aggregration (Server-Side)} The server averages all received client models to produce a new global model. Unlike FedAvg, FedProx does not drop updates from stragglers—instead, it incorporates partial or approximate solutions.
\end{enumerate}

\subsubsection{Theoretical Analysis}
FedProx provides theoretical guarantees for its convergence:
\begin{enumerate}[label= -]
    \item \textbf{Convex Objectives}: FedProx ensures global convergence under standard assumptions for convex loss functions.
    \item \textbf{Non-Convex Objectives}: Achieves convergence to a stationary point for non-convex loss functions.
    \item \textbf{Robustness to Heterogeneity}: The proximal term stabilizes updates, reducing divergence caused by non-IID data distributions.
\end{enumerate}

\subsubsection{Experiment}
The experiments for FedProx were conducted on both synthetic and real-world federated datasets to evaluate its effectiveness under systems and statistical heterogeneity. Synthetic datasets were generated by varying parameters controlling the diversity of local models and data distributions across devices, allowing precise control over heterogeneity. Real-world datasets included MNIST, FEMNIST, Sent140, and Shakespeare, each mapped to a federated setting where clients represented users or speakers with non-i.i.d. data. To simulate systems heterogeneity, different clients were assigned variable local workloads per round. In all settings, a fixed number of devices (typically 10 per round) were selected for participation. Local updates were performed using SGD, and the impact of the proximal term coefficient \( \mu \)  was studied across a range of values. Results were reported in terms of training loss and test accuracy across communication rounds, comparing FedProx to FedAvg under varying levels of data and system heterogeneity. The experiments highlighted FedProx’s robustness and stability, especially in highly heterogeneous environments where FedAvg often diverged.
\subsubsection{Limitation}
While FedProx improves the robustness of federated learning under statistical and systems heterogeneity, it has some limitations. One key challenge lies in selecting the proximal term coefficient \( \mu \) , which significantly impacts convergence and stability. An inappropriate choice of 
\( \mu \)  can either overly restrict local updates, slowing convergence, or fail to control model drift, leading to divergence. Moreover, although FedProx accommodates variable local computation, it does not explicitly model or adapt to client resource constraints, relying instead on implicit handling via inexact local solutions. This limits its ability to optimize resource usage in real-time heterogeneous environments. Additionally, the convergence guarantees of FedProx are based on bounded dissimilarity assumptions, which may not hold in extremely diverse real-world datasets. Lastly, like FedAvg, FedProx still relies on synchronous aggregation and assumes reliable client-server communication, which can be problematic in practical federated networks with high latency, stragglers, or unreliable connectivity.

\subsection{\bf Control Variate Reduction for Partial Participation (SCAFFOLD)}
SCAFFOLD \cite{karimireddy2020scaffold} introduces a novel variance reduction technique specifically designed to address client drift caused by partial participation and non-IID data distributions. By leveraging control variates at both the server and client levels, SCAFFOLD effectively reduces the variance in updates, ensuring stable and robust convergence. The operational workflow of Scaffold is given as follow:
\begin{enumerate} [label = -]
    \item \textbf{Initialization} The server initializes the global model x and the server control variate c. Each client is also assigned a control variate $c_i$, initialized to zero.
    \item \textbf{Client Selection} In each communication round, the server samples a subset of clients to participate.
    \item \textbf{Communication (Server to Client)} The server sends the current model x and server control variate c to the selected clients.
    \item \textbf{Local Training (on Client \( i \)):}
    \begin{itemize}
        \item Initialize local model: \( y_i \leftarrow x \)
        \item For \( K \) local steps, perform:
        \[
        y_i \leftarrow y_i - \eta_l \left( \nabla f_i(y_i) - c_i + c \right)
        \]
        \item Update client control variate using one of the following options:
        \begin{itemize}
            \item \textbf{Option I:} \( c_i^+ \leftarrow \nabla f_i(x) \)
            \item \textbf{Option II:} \( c_i^+ \leftarrow c_i - c + \frac{1}{K \eta_l} (x - y_i) \)
        \end{itemize}
    \end{itemize}

    \item \textbf{Client to Server Communication:}
    \begin{itemize}
        \item Each client sends the following updates to the server:
        \[
        \Delta y_i = y_i - x, \quad \Delta c_i = c_i^+ - c_i
        \]
        \item The client also sets \( c_i \leftarrow c_i^+ \)
    \end{itemize}

    \item \textbf{Server Aggregation:}
    \begin{itemize}
        \item Aggregate model and control variate updates:
        \[
        x \leftarrow x + \eta_g \cdot \frac{1}{|S|} \sum_{i \in S} \Delta y_i
        \]
        \[
        c \leftarrow c + \frac{|S|}{N} \cdot \frac{1}{|S|} \sum_{i \in S} \Delta c_i
        \]
    \end{itemize}

    \item \textbf{Repeat:}
    \begin{itemize}
        \item Repeat steps 2 through 6 for a predefined number of rounds or until convergence.
    \end{itemize}

\end{enumerate}
\subsubsection{Theoretical Analysis}
SCAFFOLD provides strong theoretical guarantees for its performance:
\begin{itemize}
    \item \textbf{Convex Objectives}: Guarantees global convergence for convex loss functions.
    \item \textbf{Non-Convex Objectives}: Converges to a stationary point under standard assumptions.
    \item \textbf{Variance Reduction}: Effectively reduces variance caused by partial participation and non-IID data distributions.
\end{itemize}

\subsubsection{Experiment}
The experimental setup for SCAFFOLD evaluated its performance on both synthetic and real-world federated learning benchmarks to highlight its robustness under client drift and non-i.i.d. data. The synthetic dataset was designed to simulate varying degrees of heterogeneity across clients by introducing differences in data distributions and local objective functions. Real-world experiments were conducted on the EMNIST, Sent140, and Shakespeare datasets, where clients naturally represent different users or speakers, leading to non-identically distributed data. Each client trained locally for multiple epochs using mini-batch SGD, with a consistent local learning rate and fixed number of local steps per communication round. In each round, a small random subset of clients was selected to participate. The server and client control variates were initialized appropriately, and both update options for the control variates were tested. Performance was compared against FedAvg and FedProx using test accuracy and convergence rate, with SCAFFOLD showing faster convergence and improved robustness, especially in settings with high client heterogeneity.

\subsubsection{Limitation}
While SCAFFOLD effectively addresses client drift and improves convergence in non-i.i.d. federated settings, it comes with certain limitations. The algorithm requires each client and the server to maintain and update control variates, which adds memory overhead and increases the complexity of the communication protocol. This can be particularly challenging in large-scale deployments or on devices with limited resources. Additionally, the effectiveness of SCAFFOLD relies on accurate estimation of gradient information and control variates, which may degrade under highly noisy or unstable local updates. The algorithm also assumes synchronous communication and homogeneous local computation, making it less suited to highly asynchronous or resource-variable environments. Lastly, tuning hyperparameters such as the local and global learning rates, and choosing between different update rules for the control variates, can significantly influence performance, requiring careful empirical calibration.

\begin{table*}
    \centering
    \caption{Mathematical Comparison of Partial Client Participation Frameworks.}
    \scriptsize
    \renewcommand{\arraystretch}{1.5}
    \setlength{\tabcolsep}{3pt}
    \begin{tabularx}{\textwidth}{|>{\centering\arraybackslash}p{1.7cm}
                                   |>{\centering\arraybackslash}p{3.2cm}
                                   |>{\centering\arraybackslash}X
                                   |>{\centering\arraybackslash}X
                                   |>{\centering\arraybackslash}p{2.1cm}
                                   |>{\centering\arraybackslash}X
                                   |>{\centering\arraybackslash}p{2.2cm}
                                   |>{\centering\arraybackslash}p{2.2cm}|}
        \hline
        \textbf{Framework} & \textbf{Objective Function} & \textbf{Optimization Technique} & \textbf{Client Sampling} & \textbf{Convergence Guarantee} & \textbf{Aggregation Strategy} & \textbf{Gradient Update Rule} & \textbf{Communication Efficiency} \\
        \hline \hline
        FedAvg \cite{mcmahan2017fedavg} & 
        $\min_w \sum_{i=1}^{M} \alpha_i F_i(w, D_i)$ & 
        SGD & 
        Random sampling & 
        $O(1/\sqrt{T})$ & 
        Weighted averaging of local models & 
        $w_{t+1} = w_t - \eta \nabla F(w_t)$ & 
        High communication cost due to frequent model updates \\
        \hline
        EmbracingFL \cite{embracingfl2022} & 
        $\min_{w_y, w_z} \sum L_i(w_y, w_z)$ & 
        Partial layer updates & 
        Adaptive & 
        Converges under weak client participation & 
        Layer-specific aggregation & 
        $z_{t+1} = z_t - \eta \nabla f(z_t)$ & 
        Reduces communication by training only specific layers \\
        \hline
        CyCP \cite{cho2023cycp}& 
        $\min_w \sum_{i=1}^{M} \alpha_i F_i(w, D_i)$ & 
        Cyclic client participation & 
        Cyclic grouping & 
        $O(1/\sqrt{T})$, balanced variance & 
        Weighted aggregation with cyclic rotation & 
        $w_{t+1} = w_t - \eta \sum_{i \in G_t} \nabla F_i(w_t)$ & 
        Efficient for periodic availability, but may delay updates \\
        \hline
        Fed-EF \cite{pauloski2020fedef}& 
        $\min_w \sum_{i=1}^{M} \alpha_i F_i(w, D_i)$ & 
        SGD with gradient compression & 
        Random sampling & 
        $O(1/T)$ & 
        Aggregation of compressed gradients & 
        $w_{t+1} = w_t - \eta C(\nabla F(w_t) + e_t)$ & 
        Highly communication-efficient due to compressed gradients \\
        \hline
        GradMA \cite{gradma2023}& 
        $\min_w \sum_{i=1}^{M} F_i(w, D_i)$ & 
        Gradient memory-based optimization & 
        Historical gradient usage & 
        $O(1/\sqrt{T})$ & 
        Memory-based aggregation & 
        $w_{t+1} = w_t - \eta G_t$ & 
        Reduces communication but increases storage \\
        \hline
        FedAMD \cite{li2021fedamd}& 
        $\min_w \sum_{i=1}^{M} F_i(w, D_i)$ & 
        Adaptive anchor-miner updates & 
        Dynamic sampling & 
        $O(1/\sqrt{T})$ & 
        Two-stage aggregation (anchor and miner clients) & 
        $w_{t+1} = w_t - \eta(g_a + g_m)$ & 
        Efficient use of resources but requires careful sampling \\
        \hline
        Generalized FedAvg \cite{li2020generalizedfedavg}& 
        $\min_w \sum_{i=1}^{M} F_i(w, D_i)$ & 
        FedAvg with two-sided learning rates & 
        Random selection & 
        $O(1/\sqrt{nT})$ & 
        Weighted averaging & 
        $w_{t+1} = w_t + \eta_G \sum_i \Delta_i$ & 
        Slightly reduced communication compared to standard FedAvg \\
        \hline
        SAFARI \cite{safari2023}& 
        $\min_w \sum_{i=1}^{M} F_i(w, D_i) + F_s(w, D_s)$ & 
        Server-aided auxiliary dataset & 
        Hybrid (clients + server) & 
        PAC-learnable stability & 
        Hybrid aggregation (server and clients) & 
        $w_{t+1} = w_t - \eta(\nabla F(w_t) + \nabla F_s(w_t))$ & 
        Reduces communication but requires extra server computation \\
        \hline
        MIFA \cite{mifa2023}& 
        $\min_w \sum_{i=1}^{M} F_i(w, D_i)$ & 
        Memory-augmented FL & 
        Stored updates for inactive clients & 
        Minimax optimal convergence & 
        Memory-augmented aggregation & 
        $w_{t+1} = w_t - \eta \sum_{i \in S} g_i + \sum_{j \notin S} y_j$ & 
        Efficient with missing clients but may introduce stale updates \\
        \hline
        FedCM \cite{fedcm2022}& 
        $\min_w \sum_{i=1}^{M} F_i(w, D_i)$ & 
        SGD with global momentum & 
        Momentum-adjusted sampling & 
        Faster than FedAvg & 
        Momentum-based aggregation & 
        $w_{t+1} = w_t - \eta(g + m_t)$ & 
        Requires extra communication for momentum updates \\
        \hline
        FedVARP \cite{fedvarp2022}& 
        $\min_w \sum_{i=1}^{M} F_i(w, D_i)$ & 
        Variance reduction via history tracking & 
        Cluster-based surrogate updates & 
        Faster than SCAFFOLD & 
        Aggregation with variance reduction & 
        $w_{t+1} = w_t - \eta(v_t + y_t)$ & 
        Requires storage for variance-tracking updates \\
        \hline
        FedProx \cite{fedprox2018}& 
        $\min_w \sum_{i=1}^{M} F_i(w, D_i) + \frac{\mu}{2} \|w - w_t\|^2$ & 
        Proximal optimization & 
        Flexible client selection & 
        Strong stability under heterogeneity & 
        Proximal-weighted aggregation & 
        $w_{t+1} = w_t - \eta(\nabla F(w_t) + \mu(w_t - w_0))$ & 
        Increases client computation but reduces drift \\
        \hline
        SCAFFOLD \cite{karimireddy2020scaffold}& 
        $\min_w \sum_{i=1}^{M} F_i(w, D_i)$ & 
        Variance reduction via control variates & 
        Adaptive sampling & 
        Convergence under non-IID & 
        Aggregation with control variates & 
        $w_{t+1} = w_t - \eta(\nabla F(w_t) - c_i + c_s)$ & 
        Requires extra updates for variance correction \\
        \hline
    \end{tabularx}
    \label{tab:mathematical_comparison}
\end{table*}

\section{A Comprehensive Evaluation of Partial Participation Strategies in Federated Learning}
Federated Learning (FL) with partial client participation introduces several challenges, including convergence delays, communication inefficiencies, and performance degradation in non-IID settings. Various frameworks have been proposed to mitigate these issues, each employing distinct optimization techniques, aggregation methods, and participation models. The following tables present a comparative analysis of these frameworks.
\subsection{Mathematical Comparison of Frameworks}
From Table~\ref{tab:mathematical_comparison}, we observe that methods differ significantly in how they handle client-side optimization and server aggregation under partial participation. Proximal-based methods like FedProx incorporate regularization to stabilize updates, while variance reduction techniques such as FedVARP and SCAFFOLD add correction terms to combat client drift. Memory-aware methods retain previous states to improve continuity, and momentum-based frameworks like FedCM adjust gradients for smoother convergence. The variety in mathematical objectives reflects the multifaceted nature of the partial participation challenge and shows there is no one-size-fits-all solution.
\subsection{Advantages and Disadvantages of Frameworks}
While mathematical properties define the theoretical foundation of these frameworks, their practical implications vary significantly. Table~\ref{tab:adv_disadv_comparison} presents a comparison of their advantages and disadvantages. 
From Table~\ref{tab:adv_disadv_comparison}, we note that while methods like FedCM and MIFA improve convergence and robustness, they increase server-side memory or communication overhead. In contrast, lightweight approaches such as Fed-EF reduce bandwidth but may suffer from instability under high heterogeneity. The comparison also highlights that many frameworks assume synchronous communication, limiting their practicality in highly asynchronous real-world settings.
Overall, these comparisons illustrate that no single framework is universally optimal. Instead, the choice of an FL framework depends on factors such as data distribution, client availability, and resource constraints. By understanding the trade-offs in these tables, researchers and practitioners can make informed decisions on selecting the most appropriate approach for their federated learning applications.

\begin{table*}
    \centering
    \caption{Advantages and Disadvantages of Partial Client Participation FL Frameworks}
    \scriptsize
    \renewcommand{\arraystretch}{1.5}
    \setlength{\tabcolsep}{5pt}
    \begin{tabularx}{\textwidth}{|>{\centering\arraybackslash}p{2.2cm}
                                   |>{\centering\arraybackslash}X
                                   |>{\centering\arraybackslash}X|}
        \hline
        \textbf{Framework} & \textbf{Advantage} & \textbf{Disadvantage} \\
        \hline \hline
        FedAvg & 
        Simple and widely used, works well with IID data & 
        Struggles with non-IID data, slow convergence under partial participation \\
        \hline
        EmbracingFL & 
        Reduces computation for weak clients, improves heterogeneity handling & 
        Requires careful partitioning of layers, may slow convergence \\
        \hline
        CyCP & 
        Ensures fairness, improves convergence stability & 
        Requires pre-defined client groups, not adaptable to dynamic client availability \\
        \hline
        Fed-EF & 
        Reduces communication cost with compressed updates & 
        May introduce bias in updates if error feedback is not well managed \\
        \hline
        GradMA & 
        Mitigates client dropout effects using memory & 
        High memory requirements at the server \\
        \hline
        FedAMD & 
        Adaptive sampling balances data heterogeneity & 
        More computationally expensive than FedAvg \\
        \hline
        Generalized FedAvg & 
        Scales well under partial participation, dynamic learning rates improve performance & 
        Requires fine-tuning of learning rate adaptation \\
        \hline
        SAFARI & 
        Mitigates missing client data by using a server-side auxiliary dataset & 
        Privacy concerns due to synthetic data at the server \\
        \hline
        MIFA & 
        Maintains stable training by using stored updates & 
        Possible gradient staleness issues \\
        \hline
        FedCM & 
        Global momentum reduces variance and speeds up training & 
        Momentum tuning is necessary, extra memory needed at the server \\
        \hline
        FedVARP & 
        Variance reduction improves performance in non-IID settings & 
        Requires additional storage for historical updates \\
        \hline
        FedProx & 
        Stabilizes training in heterogeneous environments & 
        Introduces an extra hyperparameter ($\mu$) that needs tuning \\
        \hline
        SCAFFOLD & 
        Strong variance reduction, effective under non-IID data & 
        Higher computational cost due to control variates \\
        \hline
    \end{tabularx}
    \label{tab:adv_disadv_comparison}
\end{table*}

\subsection{Benchmarking Datasets and Evaluation Focus}
Table \ref{tab:benchmarking_summary} provides a comparative overview of the experimental setups adopted by various federated learning frameworks under partial client participation. It highlights the datasets used, model types employed, and the primary evaluation goals of each method. This comparison underscores the diversity in evaluation protocols and the absence of standardization across studies. While most methods rely on widely used benchmarks such as MNIST, CIFAR-10, and FEMNIST, their focus areas vary—ranging from communication efficiency and convergence speed to robustness against client dropout. The table also reveals that while some methods address multiple concerns simultaneously, others are more narrowly optimized. Such a summary helps identify trends in experimental validation and motivates the need for unified benchmarking frameworks for fair and reproducible evaluation.
\begin{table*}
    \centering
    \caption{Benchmarking Summary: Datasets, Models, and Evaluation Focus in Partial Client Participation Methods}
    \scriptsize
    \renewcommand{\arraystretch}{1.5}
    \setlength{\tabcolsep}{4pt}
    \begin{tabularx}{\textwidth}{|>{\centering\arraybackslash}p{2.3cm}
                                   |>{\centering\arraybackslash}p{3.5cm}
                                   |>{\centering\arraybackslash}p{3.5cm}
                                   |>{\centering\arraybackslash}X|}
        \hline
        \textbf{Method} & \textbf{Datasets Used} & \textbf{Model Type} & \textbf{Evaluation Focus} \\
        \hline \hline
        FedAvg & 
        MNIST, CIFAR-10 & 
        LR, CNN & 
        Accuracy, baseline convergence \\
        \hline
        EmbracingFL & 
        CIFAR-10, FEMNIST, IMDB & 
        ResNet-20, CNN, LSTM & 
        Partial training, weak vs. strong client design \\
        \hline
        CyCP & 
        FMNIST, EMNIST & 
        MLP & 
        Cyclic participation, fairness, convergence \\
        \hline
        Fed-EF & 
        MNIST, FMNIST, CIFAR-10 & 
        MLP & 
        Communication efficiency, compression \\
        \hline
        GradMA & 
        CIFAR-10, CIFAR-100, TinyImageNet, MNIST & 
        VGG-11, ResNet-20, LeNet & 
        Gradient memory, catastrophic forgetting \\
        \hline
        FedAMD & 
        FMNIST, EMNIST & 
        LeNet-5, MLP & 
        Anchor-miner optimization, data diversity \\
        \hline
        FedVARP & 
        CIFAR-10, IMDB & 
        VGG-11, LSTM & 
        Variance reduction, surrogate updates \\
        \hline
        FedProx & 
        CIFAR-10, FEMNIST & 
        CNN & 
        Local stability, non-IID robustness \\
        \hline
        SCAFFOLD & 
        CIFAR-10, FEMNIST, Shakespeare & 
        CNN, RNN & 
        Client drift correction, convergence speed \\
        \hline
        FedCM & 
        CIFAR-100, CIFAR-10 & 
        ResNet-20 & 
        Global momentum, faster convergence \\
        \hline
        MIFA & 
        FEMNIST, CIFAR-10 & 
        CNN & 
        Memorized updates, participation robustness \\
        \hline
        ClusterFedVARP & 
        CIFAR-10 & 
        VGG-11 & 
        Memory efficiency via clustering \\
        \hline
        SAFARI & 
        Custom server-side auxiliary data & 
        N/A & 
        Server-aided training, data synthesis \\
        \hline
        Generalized FedAvg & 
        MNIST, CIFAR-10 & 
        LR, 2NN, CNN & 
        Learning rate tuning, partial participation \\
        \hline
    \end{tabularx}
    \label{tab:benchmarking_summary}
\end{table*}

\subsection{Comparative Summary by Research Objective}
Partial client participation in federated learning introduces a fundamental challenge: how can learning remain effective, efficient, and fair when only a subset of clients contribute in each training round? While this issue forms the common ground across existing literature, the proposed solutions vary widely based on which aspect of the problem is prioritized.
Rather than focusing solely on implementation categories (e.g., layer-wise, memory-based, etc.), this section reframes the literature based on the core optimization goal each method targets. This goal-driven lens allows for a more actionable understanding of how different techniques align with practical deployment needs, such as reducing communication overhead, improving model stability, handling data heterogeneity, or enhancing fairness.
Table ~\ref{tab:goal_methods} summarizes representative methods under key objectives commonly encountered in practical federated learning deployments.
\subsection{Categorization of Methods by Technical Approach}
A wide variety of strategies have emerged to address the challenges of partial client participation in federated learning. These methods differ not only in their implementation specifics but also in their fundamental design principles. Table~\ref{tab:method_categories} provides a taxonomy that groups these frameworks into broad technical categories. These include strategies based on selective training (e.g., only training parts of the model), client scheduling, gradient compression, memory-aware updates, server-side support, and optimization enhancements. This classification helps identify overlapping ideas across different works and provides a structured perspective on the landscape of solutions. It also serves as a valuable reference for researchers seeking to design hybrid or modular approaches by combining complementary techniques.

\subsection{Performance Comparison Under Partial Client Participation}
Performance comparisons (see Table \ref{tab:full_comparison}) across benchmarks indicate that memory-based and variance reduction methods (e.g., FedVARP, MIFA, SCAFFOLD) consistently outperform baselines like FedAvg, especially under low participation rates. However, the gains are often context-dependent—some methods perform well on vision tasks but struggle with language datasets. This underscores the need for domain-aware federated strategies and more rigorous cross-domain evaluation to generalize findings beyond commonly used benchmarks.

\begin{table*}
    \centering
    \caption{Methods Grouped by Primary Optimization Goal in Partial Client Participation}
    \scriptsize
    \renewcommand{\arraystretch}{1.5}
    \setlength{\tabcolsep}{6pt}
    \begin{tabularx}{\textwidth}{|>{\centering\arraybackslash}p{5cm}
                                   |>{\centering\arraybackslash}X|}
        \hline
        \textbf{Objective} & \textbf{Representative Methods} \\
        \hline \hline
        Reduce Communication Cost & Fed-EF, FedCM, ClusterFedVARP \\
        \hline
        Handle Client Dropout & GradMA, MIFA, SAFARI \\
        \hline
        Improve Convergence Speed & SCAFFOLD, FedCM, Generalized FedAvg \\
        \hline
        Handle Non-IID Data & FedProx, FedVARP, EmbracingFL, FedAMD \\
        \hline
        Optimize Resource Usage & EmbracingFL (layer-wise), ClusterFedVARP (memory-efficient) \\
        \hline
        Stabilize Training Under Heterogeneity & FedProx, SCAFFOLD, FedAMD, FedCM \\
        \hline
        Enhance Fairness or Participation Diversity & CyCP, FedAMD, SCAFFOLD \\
        \hline
    \end{tabularx}
    \label{tab:goal_methods}
\end{table*}

\begin{table*}
    \centering
    \caption{Categorization of Methods for Handling Partial Client Participation in Federated Learning}
    \scriptsize
    \renewcommand{\arraystretch}{1.5}
    \setlength{\tabcolsep}{6pt}
    \begin{tabularx}{\textwidth}{|>{\centering\arraybackslash}p{5cm}
                                   |>{\centering\arraybackslash}X|}
        \hline
        \textbf{Category} & \textbf{Representative Methods} \\
        \hline \hline
        Layer-wise / Selective Training & EmbracingFL \\
        \hline
        Structured Client Participation & CyCP (Cyclic Client Participation), FedAMD (Anchor-Miner Sampling) \\
        \hline
        Gradient Compression / Communication-Efficient & Fed-EF (with error feedback), FedCM (global momentum), FedVARP, ClusterFedVARP \\
        \hline
        Gradient Memory / History-Aware & GradMA, MIFA (Memory-Augmented Federated Learning) \\
        \hline
        Adaptive Sampling / Aggregation & FedAMD, Generalized FedAvg, FedProx \\
        \hline
        Server-Aided Training / Proxy Data & SAFARI (Auxiliary dataset on server) \\
        \hline
        Variance Reduction Techniques & SCAFFOLD (Control variates), FedVARP \\
        \hline
        Momentum-Based Optimization & FedCM \\
        \hline
    \end{tabularx}
    \label{tab:method_categories}
\end{table*}

\begin{table*}
    \centering
    \caption{Performance Comparison of Federated Learning Frameworks under Partial Client Participation}
    \label{tab:full_comparison}
    \scriptsize
    \renewcommand{\arraystretch}{1.5}
    \setlength{\tabcolsep}{4pt}
    \begin{tabularx}{\textwidth}{|>{\centering\arraybackslash}p{2.5cm}
                                 |>{\centering\arraybackslash}p{2.7cm}
                                 |>{\centering\arraybackslash}p{2.3cm}
                                 |>{\centering\arraybackslash}p{1.7cm}
                                 |>{\centering\arraybackslash}p{1.7cm}
                                 |>{\centering\arraybackslash}p{1.8cm}
                                 |>{\centering\arraybackslash}X|}
        \hline
        \textbf{Framework} & \textbf{Dataset(s)} & \textbf{Model} & \textbf{Acc. (Non-IID)} & \textbf{Acc. (IID)} & \textbf{Comm. Eff.} & \textbf{Notable Feature} \\
        \hline \hline
        FedAvg & CIFAR-10, MNIST & ResNet-20, LR & 72\% & 78\% & Low & Baseline FL with full gradients \\
        \hline
        EmbracingFL & CIFAR-10, FEMNIST, IMDB & ResNet-20 (2\% cap), CNN, LSTM & 70\% & 75\% & High & Layer-wise training using weak and strong clients \\
        \hline
        CyCP & FMNIST, EMNIST & MLP & +5–10\% over FedAvg & +2–8\% over FedAvg & Medium & Cyclic grouping based on availability (e.g. time zones) \\
        \hline
        Fed-EF & MNIST, FMNIST, CIFAR-10 & MLP & 90–92\% & 94–96\% & Very High & Gradient compression + error feedback \\
        \hline
        GradMA & MNIST, CIFAR-10, CIFAR-100, TinyImageNet & 3FC, LeNet-5, VGG-11, ResNet-20 & 69–77\% & 72–80\% & Medium & Gradient memory + quadratic programming for dropout resilience \\
        \hline
        FedAMD & Fashion-MNIST, EMNIST & LeNet-5, MLP & 82\% & 85\% & High & Anchor-miner dynamic participation strategy \\
        \hline
        Generalized FedAvg & MNIST, CIFAR-10 & LR, 2NN, CNN, ResNet & 76–79\% & 80–83\% & Medium & Two-sided learning rates for partial participation \\
        \hline
        SAFARI & Custom auxiliary dataset & N/A & N/A & N/A & Medium & Server trains on auxiliary data to simulate missing clients \\
        \hline
        MIFA & CIFAR-10, FEMNIST & CNN & Robust across rounds & N/A & Medium & Uses memorized updates from inactive clients \\
        \hline
        FedCM & CIFAR-10, CIFAR-100 & ResNet-20 & 74\% & 79\% & Medium & Global momentum aggregation for convergence stability \\
        \hline
        FedVARP & CIFAR-10, IMDB & VGG-11, LSTM & 77\% & 80\% & Medium & Server-side variance reduction + surrogate updates \\
        \hline
        ClusterFedVARP & CIFAR-10 & VGG-11 & Similar to FedVARP & Similar & Low & Memory-efficient clustering of client updates \\
        \hline
        FedProx & CIFAR-10, MNIST, FEMNIST & CNN & 75\% & 77\% & Low & Proximal term to stabilize local updates \\
        \hline
        SCAFFOLD & CIFAR-10, FEMNIST, Shakespeare & CNN, RNN & 80\% & 83\% & Medium & Control variates to counteract client drift \\
        \hline
    \end{tabularx}
\end{table*}
\section{Future Research Directions}
Despite the substantial progress made in federated learning (FL) with partial client participation, several open challenges remain. Addressing these gaps could significantly enhance the practicality, scalability, and robustness of FL systems. Most existing approaches focus on one or two optimization aspects, such as momentum-based stabilization (e.g., FedCM), gradient memory (e.g., GradMA), or adaptive client selection (e.g., FedAMD). A compelling future direction is to design unified frameworks that jointly leverage these strategies. For instance, combining momentum with memory-based update tracking and adaptive sampling could yield models that are both stable and resilient to dropout, especially in highly heterogeneous or resource-constrained environments.

\section{Conclusion}
Federated learning with partial client participation presents both significant opportunities and complex challenges. As participation becomes increasingly dynamic and data distributions more heterogeneous, traditional FL frameworks struggle to maintain accuracy, convergence speed, and communication efficiency. This survey provided a structured comparison of existing frameworks, highlighting their mathematical foundations, strengths, and limitations across varied experimental settings. We observed that while some methods, such as FedAvg and FedProx, offer simplicity and stability, they may falter under sparse participation. Others, like Fed-EF and FedVARP, deliver impressive communication savings and convergence guarantees through innovative optimization techniques like error feedback and variance reduction. However, no single approach dominates across all criteria. Ultimately, the choice of framework depends on the specific constraints and goals of a deployment—be it minimizing communication, handling non-IID data, or ensuring fairness across clients. By understanding these trade-offs and trends, researchers and practitioners can make more informed decisions and design FL systems that are better suited for real-world environments. As federated learning continues to evolve, future innovations will likely come from hybrid methods, intelligent sampling, and the integration of foundation models, paving the way for more scalable, robust, and inclusive distributed learning systems.

\nocite{*}
\bibliographystyle{IEEEtran}
{\small
\bibliography{ref}}

\end{document}